\documentclass{article}

\usepackage{microtype}
\usepackage{graphicx}
\usepackage{subfigure}
\usepackage{booktabs}
\usepackage{hyperref}

\usepackage{url}
\usepackage{graphicx}
\usepackage{subcaption}
\usepackage{multirow, multicol}
\usepackage[accepted]{icml_fmwild}
\usepackage{setspace}

\usepackage{amsmath}
\usepackage{amssymb}
\usepackage{mathtools}
\usepackage{amsthm}
\usepackage[capitalize,noabbrev]{cleveref}
\usepackage{colortbl}
\usepackage{cancel}
\usepackage{bm}
\usepackage{hyperref}
\usepackage{url}
\usepackage{graphicx}

\usepackage{adjustbox}
\usepackage{listings}
\usepackage{color}
\definecolor{codegreen}{rgb}{0,0.6,0}
\definecolor{codegray}{rgb}{0.5,0.5,0.5}
\definecolor{codepurple}{rgb}{0.58,0,0.82}
\definecolor{backcolour}{rgb}{0.95,0.95,0.92}
\definecolor{Gray}{gray}{0.9}

\theoremstyle{plain}

\theoremstyle{definition}

\theoremstyle{remark}

\icmltitlerunning{In-Context Learning Improves Compositional Understanding of Vision-Language Models}

\begin{document}

\twocolumn[
% TODO: Fill in your title.
% \icmltitle{A Study on Compositionality in Vison-Language Models}
% \icmltitle{Synthetic and Real few-shot samples improve Vison-Language Models compositionality}
% \icmltitle{Synthetic and Real Few-shot Samples Improve Compositional Understanding of Vision-Language Models}
% \icmltitle{In-Domain Synthetic and Real Image Demonstrations Improve Compositional Understanding of Vision-Language Models}
\icmltitle{In-Context Learning Improves Compositional Understanding of Vision-Language Models}
% \icmltitle{Few-shot Demonstrations Improve Compositional Understanding of Vision-Language Models}
% \icmltitle{Few-shot through CoT prompting improve VLMs compositional understanding}
% \icmltitle{Generative VLMs are few-shot compositional learners}

\begin{icmlauthorlist}
% TODO: Fill in your names here.
\icmlauthor{Matteo Nulli}{uva}
\icmlauthor{Anesa Ibrahimi}{uva}
\icmlauthor{Avik Pal}{uva}
\icmlauthor{Hoshe Lee}{uva}
\icmlauthor{Ivona Najdenkoska}{uva}
\end{icmlauthorlist}

\icmlaffiliation{uva}{University of Amsterdam, The Netherlands}
\icmlcorrespondingauthor{}{matteo.nulli@student.uva.nl}

\vskip 0.3in]

\printAffiliationsAndNotice{}

\begin{abstract}
Vision-Language Models (VLMs) have shown remarkable capabilities in a large number of downstream tasks. Nonetheless, compositional image understanding remains a rather difficult task due to the object bias present in training data. In this work, we investigate the reasons for such a lack of capability by performing an extensive bench-marking of compositional understanding in VLMs. We compare contrastive models with generative ones and analyze their differences in architecture, pre-training data, and training tasks and losses.
Furthermore, we leverage in-context learning as a way to improve the ability of VLMs to perform
more complex reasoning and understanding given an image. Our extensive experiments demonstrate that our proposed approach outperforms baseline models across multiple compositional understanding datasets. The code is available \href{https://github.com/HoeZey/vlm-compositionality/tree/main}{here}.
% Furthermore, we leverage Chain-of-Thought prompting as a way to improve the ability of VLMs to perform
% more complex reasoning and understanding given an image. Our extensive experiments demonstrate that our proposed approach outperforms baseline models across multiple compositional understanding datasets. 

% Avenues for future work on improving compositional reasoning in VLMs are discussed.  
\end{abstract}

% TODO: Comment out the guidelines.
% \input{sections/x_guidelines}

% TODO: Add more sections/files. We recommend having one file per section. Place figures in the "figures/" folder, and design tables in the "tables/" folder.

% \mn{\\Guide on commenting: I created custom commands (\textbf{ho} = Hoshee, \textbf{an} = Anesa, \textbf{mn} = Matteo, \textbf{av} = Avik) for you to write comments in the file itself. Please make an effort to use them.}

\section{Introduction}
\label{sec:intro}
Recent breakthroughs in foundation models \cite{radford2018improving, dosovitskiy2020image, BommasaniHAA21, doveh2023dac} are reaching human-level performance on many vision and language benchmarks. Particularly, Visual Language Models (VLMs) have shown impressive performance on many different downstream tasks, like image captioning \cite{yu2022coca}, visual-question answering, image understanding \cite{liu2023visual}, object localization \cite{dorkenwald2024pin} and others \cite{RadfordKHRGASAM21, SinghHGCGRK22, 0001LXH22}. Nonetheless, some tasks, that are considered simple for humans, remain hard for VLMs. Among these lies the difficulty of correctly understanding the composition of objects in an image and their textual description.
% , as depicted in Figure \ref{fig:winoground_example1}. 
% This task is defined as compositional understanding. 
% Figure \ref{fig:winoground_example1} shows examples of such complex linguistic structures connecting objects in the context of an image. 
While humans can easily connect the images correctly with their corresponding descriptions, VLMs struggle to understand this sequential order of words and thus, perform poorly \citep{yuksekgonul2023visionlanguage}. 
% Unlocking compositionality in these models remains relatively hard. 
Recent advancements in computing and data scaling resulted in huge performance increases in many VL tasks, however, \textit{compositional understanding} remains a challenging problem.
\begin{figure*}[h]
\begin{center}
\includegraphics[width=0.85\textwidth]{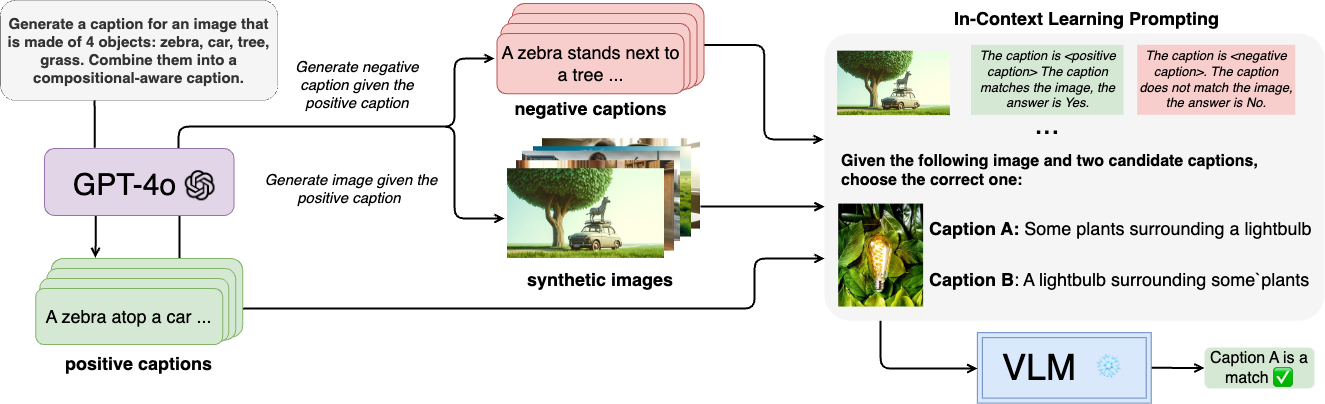}
\end{center}
\vspace{-0.5cm}
\caption{
% \textbf{Our Chain-of-Thought prompting framework for compositional understanding of VLMs.} 
\textbf{Our in-context learning pipeline for compositional understanding of VLMs.} 
We instruct GPT-4o to generate a caption consisting of a few named objects such that they are compositionally defined. Using this positive caption we then instruct GPT-4o to generate an image and a negative caption that compositionally distorts the meaning. We feed these synthetic captions and images as few-shot examples to the VLMs. Afterwards, we instruct the model to predict between correct and wrong captions for images of compositional reasoning benchmarks while also using in-context learning prompting.}
\label{fig:image_gen_pipeline}
\end{figure*}

In this work, we study how state-of-the-art VLMs perform on benchmarks that evaluate their compositional understanding \cite{thrush2022winoground, yuksekgonul2023visionlanguage, hsieh2023sugarcrepe}. Most works focus on contrastive pre-training objectives as the reason for bad performance on compositionality benchmarks \cite{yuksekgonul2023visionlanguage}. The authors reason that employing a contrastive pre-training objective pushes VLMs to perform text-image retrieval without actually having any compositional understanding. This is the main reason why such models perform well on numerous benchmarks which do not require any compositional understanding. Differently from \citet{yuksekgonul2023visionlanguage}, we evaluate both contrastive \cite{RadfordKHRGASAM21} and generative models \cite{li2023blip2, alayrac2022flamingo, liu2023visual}, to provide a deeper understanding of the underlying reasons for the lack of comprehension. Furthermore, we analyze their differences by looking at their pre-training strategy, data, and architectures. Next, we study the impact of different prompting strategies and introduce a new In-Context Learning (ICL) prompting method through synthetically generated, web images and captions. 
% Finally, we also define a new evaluation paradigm for generative models. 
Specifically, we generate the synthetic data using GPT-4o\footnote{\url{https://openai.com/index/hello-gpt-4o/}} by instructing the model to prepare compositionally aware captions from a specified list of objects. We then give this as input to GPT-4o to generate an image matching the caption and a negative caption that distorts its compositional information, as shown in Figure \ref{fig:image_gen_pipeline}. To simulate real images and captions, we randomly sample images from the COCO dataset \cite{lin2015microsoftcoco} and manually annotate a positive and a negative compositional-aware caption for them. We use these examples as demonstrations for few-shot in-context learning for the generative models along with ICL prompting.

% In summary, our contributions are the following:
% (i) We perform a comprehensive study on the behavior of generative and contrastive VLMs on several compositional understanding benchmarks.
% (ii) We introduce a CoT prompting framework by leveraging synthetic and real images and captions in a few-shot style. 
% (iii) We demonstrate the advantages of our proposed CoT framework across various few-shot settings and datasets.
In summary, our contributions are the following:
(i) We perform a comprehensive study on the behavior of generative and contrastive VLMs on several compositional understanding benchmarks.
(ii) We introduce a ICL prompting framework by leveraging synthetic and real images and captions in a few-shot style. 
(iii) We demonstrate the advantages of our proposed ICL framework for compositional image understanding across various few-shot settings and datasets.
% (iii) A different evaluation method for generative models analyzing logits outputs instead of generated text. 

\section{Related Work}
\label{sec:rel_work}

\paragraph{Compositional understanding in VLMs}
% The vast amount of large-scale pre-trained VLMs in the field of multi-modal learning has seen a surge over recent years. Advanced visual-linguistic tasks such as image-captioning have been able to be performed by these models. CLIP models \cite{RadfordKHRGASAM21}, one of the most foundational works in the field, uses large-scale image-text pairs to jointly pre-train an image encoder alongside a text encoder using a contrastive loss. Another important architecture, BLIP \cite{0001LXH22}, employs a framework leveraging captions from noisy web data for understanding and generating tasks.
Recent years have witnessed a significant increase in large-scale pre-trained VLMs within the field of multi-modal learning, enabling advanced tasks like image captioning. Foundation models such as CLIP \cite{RadfordKHRGASAM21} utilize large-scale image-text pairs for joint pre-training of image and text encoders using contrastive loss. Another notable architecture, BLIP \cite{0001LXH22}, leverages captions from noisy web data for comprehension and generation tasks. 
% The ability to grasp the underlying non-object notions of the images and text captions led to extensive research on the compositional reasoning limitations found in VLMs \citep{thrush2022winoground, yuksekgonul2023visionlanguage, doveh2023dense, hsieh2023sugarcrepe}. 
The (in)ability to grasp the underlying non-object notions of the images and text captions has led research into exploring compositional reasoning limitations of VLMs, revealing issues such as caption quality and density \citep{thrush2022winoground, yuksekgonul2023visionlanguage, doveh2023dense, hsieh2023sugarcrepe}.
%More specifically, \cite{doveh2023dense} inspected the alignment of the image-text spaces learned by VLMs and found two limiting factors, caption quality and density of captions. Output representations of VLMs were seen to behave like bags of words due to the disregarding of several object attributes appearing in texts or images \citep{yuksekgonul2023visionlanguage}. Therefore, an alternative approach contributing to the enhancement of compositional reasoning abilities of VLMs was offered via the automatic enhancement of caption quality and density of datasets \cite{urbanek2023picture}. Similarly, \cite{touvron2023llama} was introduces to propose a densely captioned images dataset, containing high-quality and dense representations ultimately proving to be useful for VLMs.
Specifically, \citet{doveh2023dense} and \citet{ yuksekgonul2023visionlanguage} found that VLMs' output representations often resemble bags of words, neglecting several object attributes. To address these limitations, enhancements in caption quality and density were proposed \citep{urbanek2023picture, touvron2023llama}, including datasets with densely captioned images, significantly improving the compositional reasoning abilities of VLMs.

\paragraph{Enhancing reasoning through In-Context Learning}
% In order to facilitate reasoning in LLMs employing chain-of-thought (CoT) prompting have been considered and shown successful capabilities \cite{wei2023chainofthought}. The decomposition of problems into intermediate steps alongside the interpretable window into the model’s behavior is what enabled the large performance gains. Recent work \cite{zhang2024cocot} builds on top of CoT prompting developing a new approach (CoCoT) which is based on the multi-input multi-modal models aimed at improving LLM’s performance in multi-image tasks. The requirement of information extraction from two images simultaneously alongside the matching of the image information with text is what distinguishes this approach from previous strategies.
% Chain-of-Thought (CoT) prompting has been employed to facilitate reasoning in Large Language Models (LLMs), demonstrating significant success by decomposing problems into intermediate steps and providing an interpretable window into the model's behavior \cite{wei2023chainofthought}. Building on CoT prompting, recent research \citet{zhang2024cocot} has developed a new approach called CoCoT, designed for multi-input multi-modal models to enhance LLMs performance in multi-image tasks. This approach requires simultaneous information extraction from two images and matching the image information with text, distinguishing it from previous strategies.
In-Context Learning (ICL) is a framework through which Large Language Models can learn a new task by being presented with demonstrations of how it should be solved.
As detailed by \citet{dong2024surveyincontextlearning} ICL has recently provided clear performance enhancements in Foundation Models, and this has come without the need of any parameter update. Indeed, ICL frameworks are easily interpretable as interacting with the model through examples is effortless and allows the model to gain understanding through comparison \cite{liu2021makes, wu2022self}.
Many studies have shown how in-context learning  capabilities can be enhanced by additionally acting on the pre-training stages of models \cite{min2021metaicl, dong2024surveyincontextlearning}. 
Regardless of the approach, ICL frameworks have shown to increase models performances on many downstream tasks and thus we set out to try its effectiveness also in compositional understanding.

\begin{figure*}[h]
\begin{center}
\includegraphics[width=1\textwidth]{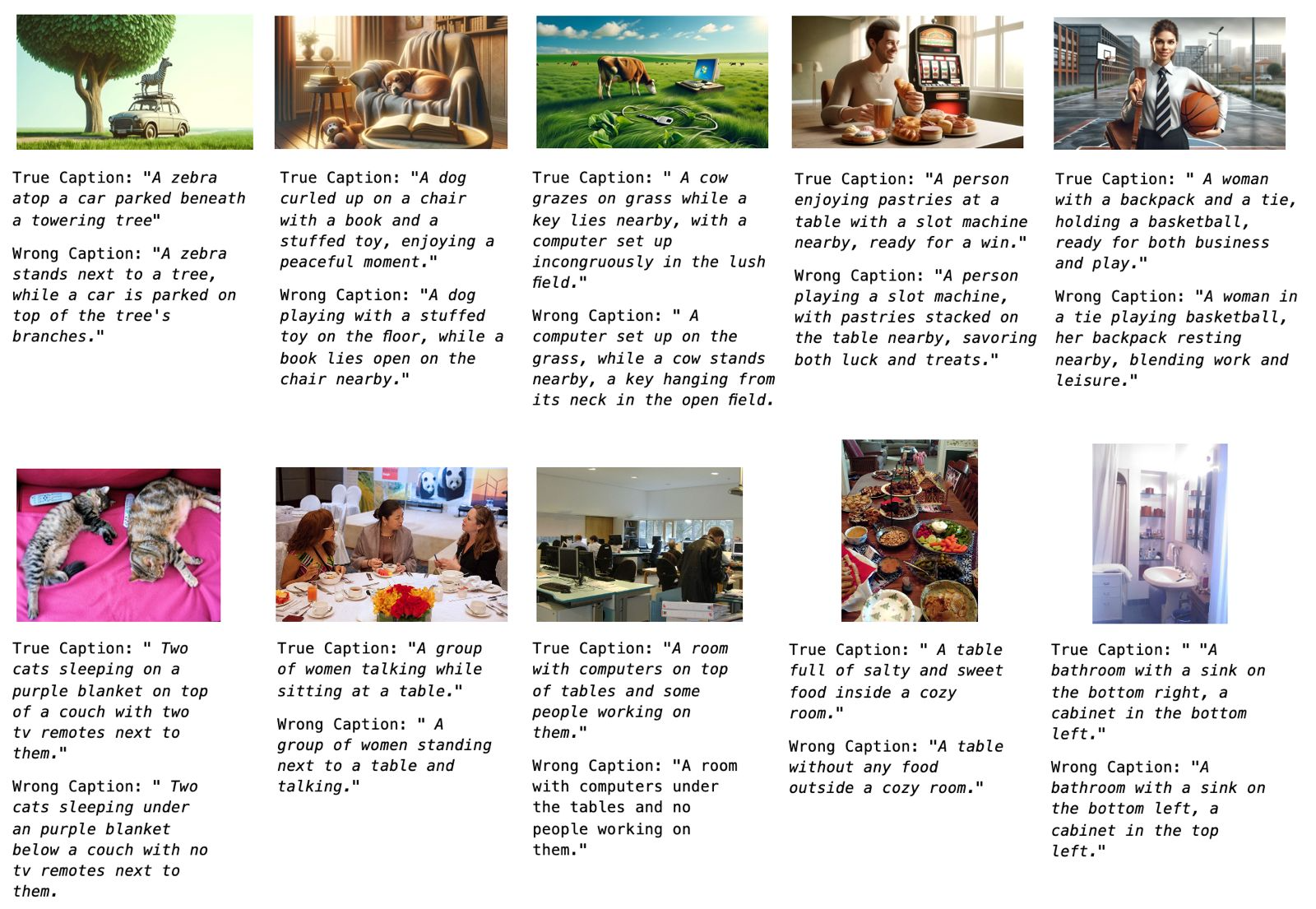}
\end{center}
\caption{\textbf{Few-shot samples}. \emph{First row}: The images and captions are synthetically generated with GPT-4o as seen in Figure \ref{fig:image_gen_pipeline}. \emph{Second row}: Images are manually captioned and retrieved from COCO dataset \cite{lin2015microsoftcoco}. We use these images to instill an understanding of the task within the generative models in a few-shot manner.}
\label{fig:few_shot_examples}
\end{figure*}

\section{Methodology} 
% Different prompting strategies have proven to be effective in increasing model capabilities during inference \cite{brown2020language, zhang2023multimodal, kojima2023large}. 
This paper explores how in-context learning prompts consisting of intermediate reasoning steps can help VLMs to be more compositionally aware. To that end, we introduce an in-context learning prompting framework, presented in Figure \ref{fig:image_gen_pipeline}. The intermediate steps consist of images and corresponding captions in a few-shot style. We hypothesize that by doing this, VLMs can understand the complex compositional relations within an image through correct and incorrect examples of image-caption pairs. In the following sections we will describe the process in details.

\paragraph{Few-shot ICL prompting with synthetic images} \label{par: synth}
% We propose a ICL prompting strategy through the help of few-shot synthetic image examples as shown in Figure \ref{fig:image_gen_pipeline}. 
First, we generate a compositional aware caption with GPT-4o by prompting it in the following way:\\
\textit{Generate a caption for an image which is made of 4 objects: object 1, object 2, object 3, object 4. Can you combine them into a compositionally aware caption?}\\
This will return the \textit{positive} caption which is then fed back into GPT-4o, to generate a corresponding image with 
\href{https://openai.com/index/dall-e-3/}{DALL·E}.
% DALL·E\footnote{\url{https://openai.com/index/dall-e-3/}}. 
Finally, with GPT-4o, we also generate a compositionally-aware \textit{negative} caption, that is in contrast with the correct caption, using the prompt of Appendix \ref{app: method}.
% This shows the model a compositional-aware hard-negative example for each image. 
The same procedure is repeated five times and these examples serve as few-shot examples for the model to understand the compositional nature of the images. The first row of Figure \ref{fig:few_shot_examples} shows the outputs of the described generation step. Along with these 5 examples, the model is also given a query input, as shown in Appendix \ref{app: method}. 
% \textit{CoT Prompt: "Given this image and two candidate captions (A and B), which caption is the better description of the given image? Think step-by-step and analyze each caption against the image. Begin by describing the key elements visible in the image. Then, compare these elements with the details mentioned in each caption to determine which one matches better. After providing a detailed explanation of your reasoning, clearly state your final answer as $<A>$ or $<B>$."}
This helps the model reason about its generation and clearly understand the task. Additionally, we randomly switch the correct caption for few-shot samples between \textit{A} and \textit{B}, to remove any bias toward choosing either of the characters, especially since \textit{A} itself is a common token. Depending on the benchmark we use slight variations of the above prompt. More information is provided in Appendix \ref{app: pipeline}.

\paragraph{Few-shot ICL prompting with real images}
We also explore ICL prompting by employing real images. We use five images from COCO dataset \cite{lin2015microsoftcoco} and caption them manually with a positive and negative caption, to be as compositionally-aware as possible. Similar to synthetic images, we use these image-caption pairs as few-shot examples with the same prompting template. From the \emph{second row} of Figure \ref{fig:few_shot_examples} it can be seen that real ones provide a more complex and noisier context unlike synthetic images.

\begin{table*}[t!]
    \centering
    \small
    \setlength\aboverulesep{0pt}
    \setlength\belowrulesep{0pt}
    \setstretch{1.2}
    \begin{adjustbox}{center}
    \begin{tabular}{l|ccc|ccccccc|cccc|c}\toprule
    \multirow{2}{*}{\textbf{Model}} & \multicolumn{3}{c|}{\textbf{Winoground}} & \multicolumn{7}{c|}{\textbf{SugarCrepe}} & \multicolumn{4}{c|}{\textbf{ARO}} & \multirow{2}{*}{\textbf{Avg.}}\\
    \cmidrule(lr){2-4} \cmidrule(lr){5-11}\cmidrule(lr){12-15}
       & T & I & G & AO & AA & RA & RO & RR & SA & SO & C & F & VG-A & VG-R\\ \midrule
    
    \rowcolor{Gray}\multicolumn{3}{l}{\textbf{Contrastive Models}} & & & & & & & & & & & & &\\
    ViT-B-16 & 32.7 & 12.5 & 10.0 &                  % Winoground
    89.2 & 78.0 & 85.9 & 95.1 & 69.0 & 67.5 & 63.0 & % SugarCrepe 
    18.4 & 34.3 & 57.4 & 22.1 &                    % ARO
    52.5 \\
    
    ViT-B-32 & 34.3 & 10.8 & 7.5 &                     % Winoground
    87.1 & 77.9 & 82.6 & 93.8 & 68.9 & 67.4 & 60.2 & % SugarCrepe 
    32.6 & 39.5 & 61.8 & 46.1 & {55.0} \\ % ARO
    
    bigG-CLIPA$^{*1}$ & 34.0 & 13.0 & 10.3 &                    % Winoground
    \textbf{90.9} & \textbf{85.5} & {84.3} & \textbf{96.9} & 71.1 & 75.2 & 62.6 & % SugarCrepe 
    30.4 & 33.9 & 59.3 & 38.8 &                     % ARO
    56.2 \\
    
    SigLIP-256$^{*2}$ & \underline{36.7} & \underline{15.2} & \underline{13.5} & % Winoground
    \underline{90.5} & 82.9 & 85.7 & 96.1 & 69.9 & \underline{77.0} & 68.2 & % SugarCrepe 
    33.7 & 40.0 & 62.9 & 37.7 & 57.9\\                   % ARO
    
    SO-SigLIP$^{*3}$ & \textbf{38.0} & \textbf{20.3} & \textbf{15.8} &                     % Winoground
    84.0 & \underline{84.6} & 82.8 & 96.1 & 73.3 & 76.8 & 61.3 & % SugarCrepe 
    37.9 & 40.3 & \underline{65.1} & 39.1 & 58.2 \\   % ARO

    \rowcolor{Gray}\multicolumn{3}{l}{\textbf{Generative Models}} & & & & & & & & & & & & &\\

    LLaVA & 20.1 & 1.00 & 0.01 &                     % Winoground
    {78.1} & {62.8} & \underline{89.7} & {94.1} & \underline{81.2} & {69.3} & \textbf{71.9} & % SugarCrepe 
    \underline{78.9} & 83.4 & 61.0 & 54.1 & \underline{60.4} \\ % ARO
    
    CogVLM & 32.3 & {1.00} & {0.70} & % Winoground
    {79.7} & {74.2} & \textbf{91.4} & \underline{96.7} & \textbf{82.5} & \textbf{82.8} & \underline{71.5} & % SugarCrepe 
    \textbf{84.9} & \textbf{87.5} & \textbf{74.1} & \textbf{67.5} & % ARO
    \textbf{66.2} \\                     
    
    \bottomrule
    \end{tabular}
    \end{adjustbox}
    \vspace{-0.3cm}

    \caption{
    \textbf{Zero-shot performance of contrastive and generative models} on all the benchmarks and accompanying subscores. 
    *Full Model names - 1: ViT-bigG-14-CLIPA, 2: ViT-L-16-SigLIP-256, 3: ViT-SO400M-14-SigLIP.
    \textbf{Winoground}: T=text, I=Image, G=Group. 
    \textbf{SugarCrepe}: AO=add\_obj, AA=add\_att, RA=repl\_att, RO=repl\_obj, RR=repl\_rel, SA=swap\_att, SO=swap\_obj. 
    \textbf{ARO}: C=COCO, F=Flickr30k, VG-A=VG-Attribute, VG-R=VG-Relation.
    }
    \label{tab:benchmark_results_zero_shot}
\end{table*}
\vspace{-0.2cm}
\section{Experiments \& Results}
\label{sec:results}

% \subsection{Compositional Reasoning Benchmarking}

\subsection{Datasets }
\label{sec:setup}
We employ the following three datasets for evaluation of compositional understanding.
\paragraph{Attribution, Relation, and Order (ARO)} \cite{yuksekgonul2023visionlanguage} evaluates four different aspects of compositionality. \emph{Visual Genome Attributions} and \emph{Visual Genome Relations} are testing the model's understanding of correct attributes and relations associated with objects within an image, and COCO Order and Flickr30k Order are testing the understanding of the correct ordering of words in a caption.
\paragraph{Winoground} \cite{thrush2022winoground}, comprising an evaluation set of 400 pairs of two images and two captions. The pairs are very similar to each other, with slight linguistic differences in the captions.
\paragraph{SugarCrepe} \cite{hsieh2023sugarcrepe} benchmark tests various fine-grained compositional concept understanding aspects using COCO image-text pairs. An object, attribute, or relation is either replaced, swapped, or added to the original text such that the caption no longer matches the scene. 
\vspace{-0.2cm}
\subsection{Baseline models}
% In this section, we discuss the process of choosing the models to evaluate, their differences, and the type of prompting strategies to adopt based on the situation. 
% We carry out our experiments on two categories of VLMs, contrastive models from the CLIP family \cite{RadfordKHRGASAM21} and generative models \cite{alayrac2022flamingo}.

\textbf{Contrastive Models}\,\,\,\,
Most state-of-the-art VLMs are trained with a contrastive loss  \cite{RadfordKHRGASAM21, chen2020simple, zhai2023sigmoid, sun2023evaclip, li2023inverse}, and most of these use a Vision Transformer (ViT) \cite{dosovitskiy2020image} as vision encoders, while some employ ResNet architecture \cite{HeZRS16}. 
In our analysis, we use CLIP-based models from the \href{https://github.com/mlfoundations/open_clip}{OpenCLIP library}
% OpenCLIP library\footnote{\url{https://github.com/mlfoundations/open_clip}} 
\cite{ilharco_gabriel_2021_5143773}. 
Some of them include EvaCLIP \cite{sun2023evaclip}, SigLIP \cite{zhai2023sigmoid}, CLIPA \cite{li2023inverse}, and CoCa \cite{yu2022coca}.

\textbf{Generative Models}\,\,\,\,
Generative models differ from CLIP-based models, not only in their ability to perform auto-regressive generation but also in their training process.
In Section \ref{sec:results} we analyze two generative models, namely LLaVA \cite{liu2023visual} and CogVLM \cite{wang2024cogvlm}. 
 % \href{https://arxiv.org/abs/2202.12837}{re-thinkging role}
% How we choose the models, how we choose the prompting strategies?

% \subsection{Evaluation pipelines}
% We explain the evaluation pipeline employed to obtain the results of both contrastive and generative models in this section.

\begin{table*}[t!]
    \centering
    \setlength\aboverulesep{0pt}
    \setlength\belowrulesep{0pt}
    \setstretch{1.2}    
    \small
    \resizebox{\textwidth}{!}{%
    \begin{tabular}{l|l|ccc|ccccccc|cccc}\toprule
    \multirow{2}{4em}{\textbf{Method}} & \multirow{2}{4em}{\textbf{Type of\\samples}} &\multicolumn{3}{c|}{\textbf{Winoground}} & \multicolumn{7}{c|}{\textbf{SugarCrepe}} & \multicolumn{4}{c}{\textbf{ARO}} %& \multirow{2}{*}{\textbf{Avg.}}
    \\
    \cmidrule(lr){3-5} \cmidrule(lr){6-12}\cmidrule(lr){13-16}
    & & T & I & G & AO & AA & RA & RO & RR & SA & SO & C & F & VG-A & VG-R \\
    \midrule
    % \rowcolor{Gray}\multicolumn{14}{l}{\textbf{Zero-Shot}} \\
    Zero-Shot & - & 20.1 & 1.00 & 0.01 &                     % Winoground
    \textbf{78.1} & \textbf{62.8} & \textbf{89.7} & \textbf{94.1} & \textbf{81.2} & \textbf{69.3} & \textbf{71.9} & % SugarCrepe 
    78.9 & 83.4 & 61.0 & 54.1 %& 60.4 % ARO                 
    \\
        \hline
    
    % \rowcolor{Gray}\multicolumn{14}{l}{\textbf{Synthetic 1-Shot}} \\
    \multirow{2}{4em}{1-Shot} & Synthetic & 21.0 & 2.00 & 0.70 & % Winoground
    71.1 & 61.9 & 77.7 & 89.5 & 73.2 & 64.7 & 59.3 & % SugarCrepe
    79.3 & 85.3 & 58.8 & 58.7 %& 57.4 % ARO
    \\
    
    % \hline
    % \rowcolor{Gray}\multicolumn{14}{l}{\textbf{Real 1-Shot}}\\
    & Real & \textbf{26.0} & \textbf{5.70} & \textbf{2.10} & % Winoground
    72.9 & \underline{61.1} & 75.6 & 89.5 & 73.3 & 65.0 & 60.9 & % SugarCrepe
    81.8 & 84.9 & \underline{63.6} & 61.5 %& 58.9 % ARO 
    \\
    
    \hline
    % \rowcolor{Gray}\multicolumn{14}{l}{\textbf{Synthetic 5-Shot}}\\
    \multirow{2}{4em}{5-Shot} & Synthetic & 25.0 & \underline{5.00}  & \underline{2.00} & %Winoground
     \underline{75.8} & 58.9 & 81.2 &  91.2& 80.1 & \underline{65.4} & 59.7 & % SugarCrepe
    \textbf{85.8} & \underline{88.8} & \textbf{65.7} & \textbf{65.3}% & 60.7 % ARO
    \\

    % \hline
    % \rowcolor{Gray}\multicolumn{14}{l}{\textbf{Real 5-Shot}}\\
    & Real & \underline{25.5} & 4.70 & \textbf{2.10} & % Winoground
    74.7 & 60.2 & \underline{84.0} & \underline{92.7} & \underline{80.5} & \underline{65.4} & \underline{61.7} & % SugarCrepe
    \underline{85.7} & \textbf{89.1} & 62.7 & \underline{63.1} % ARO
    %\underline{60.9} 
    \\
    
    \bottomrule
    \end{tabular}
    }
    \vspace{-0.4cm}

    \caption{
    \textbf{Performance of LLaVA in a zero-shot and our in-context learning setting} using synthetic and real image and caption demonstrations. Synthetic demonstrations are generated as seen in Figure \ref{fig:image_gen_pipeline} and real images are taken from COCO and manually annotated.
    The meaning of each sub-score follows that of Table \ref{tab:benchmark_results_zero_shot}.}
    
    \label{tab:benchmark_results_few_shot}
\end{table*}
\vspace{-0.2cm}
\subsection{Evaluation with contrastive VLMs}

% The contrastive models show high performance on a wide range of downstream tasks. Thus, exploring their level of compositional understanding is beneficial to grasp the state of SOTA models. 
% For contrastive evaluation, we choose models from the OpenClip library \cite{ilharco_gabriel_2021_5143773}. We evaluate all available models and in Table \ref{tab:benchmark_results_zero_shot} we extract the top 5 models with best performance. 
% The difference among these models is in their pre-trained visual encoder. 
For contrastive evaluation, we select models from the OpenCLIP library \cite{ilharco_gabriel_2021_5143773} and assess all available options. Table \ref{tab:benchmark_results_zero_shot} highlights the top five models with the best performance, distinguished by their pre-trained visual encoders.
Table \ref{tab:benchmark_results_zero_shot} shows a general pattern in the consistent out-performance of SigLIP models (ViT-SO400M-14-SigLIP, ViT-L-16-SigLIP-256) across most benchmarks and sub-scores with an average increase of 4\% on Winoground and 6\% on ARO. This is particularly relevant in tasks requiring the replacement or swapping of attributes/objects. The ViT-bigG-CLIPA model also performs competitively, especially on SugarCrepe where it achieves a slight increase concerning SigLIP.
% Consistently, SigLIP models (ViT-SO400M-14-SigLIP, ViT-L-16-SigLIP-256) outperformed others across most benchmarks and sub-scores, showing an average performance increase of 4 on Winoground and 6 on ARO, particularly excelling in tasks involving attribute or object replacement. Additionally, the ViT-bigG-CLIPA model demonstrated competitive performance, especially on SugarCrepe, where it slightly surpassed SigLIP.
% As emphasized in \cite{RadfordKHRGASAM21}, the distinctive feature found in CLIP models is the joint training of an image and text encoder to minimize embedding distances of image-text pairings of training examples. During the testing phase, the text encoder utilizes a zero-shot linear classifier by embedding descriptions of the target dataset’s classes. 
The efficiency and ability of CLIP to perform on a wide range of tasks has been covered in \citet{RadfordKHRGASAM21}, as well as its limitations. One example is its poor performance on various fine-grained classification tasks that involve differentiating between different representations of objects. 
% Another drawback of CLIP is its poor generalizability to out-of-distribution data. 
However, one of the most significant limitations is the restriction of only being able to choose among concepts from a given zero-shot classifier. This in effect, prevents it from generating novel outputs or combining existing concepts in new ways. Consequently, when a new instance is presented, CLIP is unable to accurately classify or generate a novel output which is a key component of compositional reasoning. 
Differently, the superior performance of SigLIP models over CLIPA is also attributed to the loss and training used for SigLIP. Indeed, CLIPA is using softmax normalization in the contrastive loss which therefore normalizes every positive pair with all negative ones leading to quadratic complexity. On the contrary, SigLIP reduces the calculation to a simpler sigmoid function and independently evaluates the positive-negative pairs in the batch. This allows SigLIP to be trained more efficiently and perform better at small batch sizes.
% Other extensive analyses \cite{yuksekgonul2023visionlanguage, doveh2023dense, thrush2022winoground} have also showcased the lack of compositional reasoning and understanding in VLMs such as CLIP. By generating two novel datasets for exploring relation and attribution understanding, \cite{yuksekgonul2023visionlanguage} exhibited the deficiencies of VLMs in relational understanding. Furthermore, CLIP’s inability to encode order was also noted and considered to be a contributing factor to its struggle in generating images to the relations in the descriptions when used as the text encoder \cite{kamath2023text}. These limitations have been hypothesized by \cite{yuksekgonul2023visionlanguage} to have arisen from the way VLMs, such as CLIP, are trained.  More precisely, the authors correlate the usage of contrastive loss with VLM's inability to learn compositional reasoning. This is due to the neglect of information order in various retrieval tasks demonstrated by VLMs. Consequently, showing how models tend to adopt a shortcut strategy (bag-of-words), which ignores order information but still succeeds due to the dataset's design.
\vspace{-0.2cm}
\subsection{Evaluation with generative VLMs}

\paragraph{Zero-shot performance}
% CogVLM > LLaVA
%  - CogVLM architecture + training

We demonstrate the zero-shot performance of generative models in Table \ref{tab:benchmark_results_zero_shot}. It can be seen that CogVLM performs better than LLaVA, by increasing the Winoground text score by almost 12\% and being consistently better with an average increase of 8\% on ARO and similar yet overall better scores on SugarCrepe.
The primary reason for this is its architectural and training advantages. CogVLM employs a vision expert module at each layer of the Large Language Model (LLM) comprising of new \emph{Query-Key-Value} and MLP weights initialized from the LLM. These are tuned for vision features while the original weights for the text remain frozen, allowing for using the already learned semantics of the LLM to make better use of image features. LLaVA instead makes simpler architectural choices tuning the LLM for image and text features together. Additionally, CogVLM uses both next-token prediction and object localization in its training, whereas LLaVA only uses next-token prediction. 
% Directly optimizing for object localization can improve the model's ability to understand aspects of image composition. This does however not necessarily convey an understanding of object relations. While outside the scope of this work, future work could perform ablation studies for architectural and training decisions to see their impact on compositional understanding.

\textbf{Comparison to contrastive models}\,\,\,\,
Text encoders of CLIP-like models are order-agnostic \cite{yuksekgonul2023visionlanguage} and consequently do not perform well on the COCO-Order and Flickr30k-Order datasets of ARO. In contrast, generative models perform quite well on these tasks as the LLMs are pre-trained in a next-token prediction fashion. 
% quite sensitive to the semantic differences resulting from word orderings due to their excellent language understanding capabilities from being exposed to large text datasets. 
Regarding Winoground, generative VLMs show somewhat comparable performance on the text subscore but show quite degraded performance on image (and consequently group) scores. This could be explained by how these different model types match images and texts. Contrastive models match images and text by calculating the similarity of their logits, which is commutative, i.e. there is explicit direction from image to text or text to image. Generative VLMs on the other hand are commonly trained using image-captioning objectives and therefore have an explicit direction from image to text (i.e., describe the image shown to it) and thus, struggle when having to do the non-descriptive task of choosing between images given a caption. 
One could remedy this by comparing the caption sequence probability conditioned on different image inputs and matching based on these probabilities, but whether that truly reflects compositional understanding is unclear.
% Regarding Winoground, generative VLMs perform comparably on text subscores but poorly on image and group scores due to their training method. Contrastive models, which calculate image-text similarity in a commutative way, outperform generative models, which use autoregressive training and focus on describing images. Generative VLMs struggle with tasks requiring non-descriptive image selection based on captions. An alternative approach could involve comparing caption probabilities conditioned on different images, though its effectiveness in reflecting true compositional understanding is uncertain.

% Preliminary results show that...

\textbf{Few-shot performance}\,\,\,\,
In Table \ref{tab:benchmark_results_few_shot}, we observe that both synthetic and real demonstrations improve the performance on Winoground and ARO, but decrease on SugarCrepe.
One reason for this could be that the way we generate negative captions for both synthetic and real image demonstrations might not lend itself to SugarCrepe, leading to out-of-domain image-caption correspondences. 
Specifically, each sub-experiment within SugarCrepe creates negative captions by changing one or two aspects of the positive caption by adding, swapping, or replacing, objects or their attributes and relations. The negative captions of our demonstrations change multiple of these aspects at once. This style, however, aligns much closer to how positive and negative captions are used in Winoground and ARO, explaining the discrepancy between these benchmarks and SugarCrepe.
% Furthermore, we see that one-shot performance for real images in some instances outperforms 5-shot performance (Winoground, SugarCrepe AA, ARO VG-A). As we did not vary the image used for the 1-shot performance (we chose the left-most image for both synthetic and real images in Figure \ref{fig:few_shot_examples}), the increase in performance could be demonstration-specific, and a result of the chosen demonstration being a better fit for the given sub-experiment of the benchmark.

.

\section{Conclusion}
\label{sec:conclusion}

In this work, we explore the compositional understanding of contrastive and generative VLMs.
Despite lower language understanding, contrastive models remain competitive due to their consistent evaluation method. Generative models face challenges such as asymmetric text-image relationships due to autoregressive training and reliance on frozen CLIP-like vision encoders. 
Furthermore, we introduce an ICL framework to examine the impact of synthetic and real images and captions as few-shot demonstrations. Our results show improved performance across diverse compositional understanding benchmarks, both when using synthetic and real images. This suggests potential benefits from using task-specific, few-shot examples for improving the capabilities of VLMs, such as compositional understanding. 

\textbf{Future work}\,\,\,\,
To improve compositional understanding, future VLMs could move away from contrastive vision encoders and make use of alternative training objective like patch-level prediction \cite{oquab2024dinov2learningrobustvisual, yun2022patchpred} which has shown improved inter-patch understanding which could be useful for compositional understanding. To achieve similar results Densely Captioned Images have shown positive impact on compositional understanding \cite{urbanek2024pictureworth77text}, with further research possibly leading to substantial improvements. 
Alternatively, as compositional reasoning can be seen as a form of symbolic reasoning, transformer-based foundation models could be supplemented with logic components. Indeed, recent work has used neurosymbolic grounding to enable compositionally aware world models \cite{sehgal2024neurosymbolic}. As such, improving compositionality could be seen as falling under the larger umbrella of improving the reasoning capabilities of (multi-modal) foundation models, which might require more explicit symbolic components or finding non-symbolic architectures that can exhibit stronger machine cognition characteristics.

\bibliography{references}

\begin{thebibliography}{34}
\providecommand{\natexlab}[1]{#1}
\providecommand{\url}[1]{\texttt{#1}}
\expandafter\ifx\csname urlstyle\endcsname\relax
  \providecommand{\doi}[1]{doi: #1}\else
  \providecommand{\doi}{doi: \begingroup \urlstyle{rm}\Url}\fi

\bibitem[Alayrac et~al.(2022)Alayrac, Donahue, Luc, Miech, Barr, Hasson, Lenc, Mensch, Millican, Reynolds, Ring, Rutherford, Cabi, Han, Gong, Samangooei, Monteiro, Menick, Borgeaud, Brock, Nematzadeh, Sharifzadeh, Binkowski, Barreira, Vinyals, Zisserman, and Simonyan]{alayrac2022flamingo}
Alayrac, J., Donahue, J., Luc, P., Miech, A., Barr, I., Hasson, Y., Lenc, K., Mensch, A., Millican, K., Reynolds, M., Ring, R., Rutherford, E., Cabi, S., Han, T., Gong, Z., Samangooei, S., Monteiro, M., Menick, J.~L., Borgeaud, S., Brock, A., Nematzadeh, A., Sharifzadeh, S., Binkowski, M., Barreira, R., Vinyals, O., Zisserman, A., and Simonyan, K.
\newblock Flamingo: a visual language model for few-shot learning.
\newblock In Koyejo, S., Mohamed, S., Agarwal, A., Belgrave, D., Cho, K., and Oh, A. (eds.), \emph{Advances in Neural Information Processing Systems 35: Annual Conference on Neural Information Processing Systems 2022, NeurIPS 2022, New Orleans, LA, USA, November 28 - December 9, 2022}, 2022.

\bibitem[Bommasani et~al.(2021)Bommasani, Hudson, Adeli, Altman, Arora, von Arx, Bernstein, Bohg, Bosselut, Brunskill, Brynjolfsson, Buch, Card, Castellon, Chatterji, Chen, Creel, Davis, Demszky, Donahue, Doumbouya, Durmus, Ermon, Etchemendy, Ethayarajh, Fei{-}Fei, Finn, Gale, Gillespie, Goel, Goodman, Grossman, Guha, Hashimoto, Henderson, Hewitt, Ho, Hong, Hsu, Huang, Icard, Jain, Jurafsky, Kalluri, Karamcheti, Keeling, Khani, Khattab, Koh, Krass, Krishna, Kuditipudi, and et~al.]{BommasaniHAA21}
Bommasani, R., Hudson, D.~A., Adeli, E., Altman, R.~B., Arora, S., von Arx, S., Bernstein, M.~S., Bohg, J., Bosselut, A., Brunskill, E., Brynjolfsson, E., Buch, S., Card, D., Castellon, R., Chatterji, N.~S., Chen, A.~S., Creel, K., Davis, J.~Q., Demszky, D., Donahue, C., Doumbouya, M., Durmus, E., Ermon, S., Etchemendy, J., Ethayarajh, K., Fei{-}Fei, L., Finn, C., Gale, T., Gillespie, L., Goel, K., Goodman, N.~D., Grossman, S., Guha, N., Hashimoto, T., Henderson, P., Hewitt, J., Ho, D.~E., Hong, J., Hsu, K., Huang, J., Icard, T., Jain, S., Jurafsky, D., Kalluri, P., Karamcheti, S., Keeling, G., Khani, F., Khattab, O., Koh, P.~W., Krass, M.~S., Krishna, R., Kuditipudi, R., and et~al.
\newblock On the opportunities and risks of foundation models.
\newblock \emph{CoRR}, abs/2108.07258, 2021.

\bibitem[Chen et~al.(2020)Chen, Kornblith, Norouzi, and Hinton]{chen2020simple}
Chen, T., Kornblith, S., Norouzi, M., and Hinton, G.~E.
\newblock A simple framework for contrastive learning of visual representations.
\newblock In \emph{Proceedings of the 37th International Conference on Machine Learning, {ICML} 2020, 13-18 July 2020, Virtual Event}, volume 119 of \emph{Proceedings of Machine Learning Research}, pp.\  1597--1607. {PMLR}, 2020.

\bibitem[Dong et~al.(2024)Dong, Li, Dai, Zheng, Ma, Li, Xia, Xu, Wu, Chang, Sun, Li, and Sui]{dong2024surveyincontextlearning}
Dong, Q., Li, L., Dai, D., Zheng, C., Ma, J., Li, R., Xia, H., Xu, J., Wu, Z., Chang, B., Sun, X., Li, L., and Sui, Z.
\newblock A survey on in-context learning, 2024.
\newblock URL \url{https://arxiv.org/abs/2301.00234}.

\bibitem[Dorkenwald et~al.(2024)Dorkenwald, Barazani, Snoek, and Asano]{dorkenwald2024pin}
Dorkenwald, M., Barazani, N., Snoek, C. G.~M., and Asano, Y.~M.
\newblock {PIN:} positional insert unlocks object localisation abilities in vlms.
\newblock \emph{CoRR}, abs/2402.08657, 2024.
\newblock \doi{10.48550/ARXIV.2402.08657}.

\bibitem[Dosovitskiy et~al.(2021)Dosovitskiy, Beyer, Kolesnikov, Weissenborn, Zhai, Unterthiner, Dehghani, Minderer, Heigold, Gelly, Uszkoreit, and Houlsby]{dosovitskiy2020image}
Dosovitskiy, A., Beyer, L., Kolesnikov, A., Weissenborn, D., Zhai, X., Unterthiner, T., Dehghani, M., Minderer, M., Heigold, G., Gelly, S., Uszkoreit, J., and Houlsby, N.
\newblock An image is worth 16x16 words: Transformers for image recognition at scale.
\newblock In \emph{9th International Conference on Learning Representations, {ICLR} 2021, Virtual Event, Austria, May 3-7, 2021}. OpenReview.net, 2021.

\bibitem[Doveh et~al.(2023{\natexlab{a}})Doveh, Arbelle, Harary, Herzig, Kim, Cascante-bonilla, Alfassy, Panda, Giryes, Feris, Ullman, and Karlinsky]{doveh2023dac}
Doveh, S., Arbelle, A., Harary, S., Herzig, R., Kim, D., Cascante-bonilla, P., Alfassy, A., Panda, R., Giryes, R., Feris, R., Ullman, S., and Karlinsky, L.
\newblock Dense and aligned captions (dac) promote compositional reasoning in vl models, 2023{\natexlab{a}}.

\bibitem[Doveh et~al.(2023{\natexlab{b}})Doveh, Arbelle, Harary, Herzig, Kim, Cascante{-}Bonilla, Alfassy, Panda, Giryes, Feris, Ullman, and Karlinsky]{doveh2023dense}
Doveh, S., Arbelle, A., Harary, S., Herzig, R., Kim, D., Cascante{-}Bonilla, P., Alfassy, A., Panda, R., Giryes, R., Feris, R., Ullman, S., and Karlinsky, L.
\newblock Dense and aligned captions {(DAC)} promote compositional reasoning in {VL} models.
\newblock In Oh, A., Naumann, T., Globerson, A., Saenko, K., Hardt, M., and Levine, S. (eds.), \emph{Advances in Neural Information Processing Systems 36: Annual Conference on Neural Information Processing Systems 2023, NeurIPS 2023, New Orleans, LA, USA, December 10 - 16, 2023}, 2023{\natexlab{b}}.

\bibitem[He et~al.(2016)He, Zhang, Ren, and Sun]{HeZRS16}
He, K., Zhang, X., Ren, S., and Sun, J.
\newblock Deep residual learning for image recognition.
\newblock In \emph{2016 {IEEE} Conference on Computer Vision and Pattern Recognition, {CVPR} 2016, Las Vegas, NV, USA, June 27-30, 2016}, pp.\  770--778. {IEEE} Computer Society, 2016.
\newblock \doi{10.1109/CVPR.2016.90}.

\bibitem[Hsieh et~al.(2023)Hsieh, Zhang, Ma, Kembhavi, and Krishna]{hsieh2023sugarcrepe}
Hsieh, C., Zhang, J., Ma, Z., Kembhavi, A., and Krishna, R.
\newblock Sugarcrepe: Fixing hackable benchmarks for vision-language compositionality.
\newblock In Oh, A., Naumann, T., Globerson, A., Saenko, K., Hardt, M., and Levine, S. (eds.), \emph{Advances in Neural Information Processing Systems 36: Annual Conference on Neural Information Processing Systems 2023, NeurIPS 2023, New Orleans, LA, USA, December 10 - 16, 2023}, 2023.

\bibitem[Ilharco et~al.(2021)Ilharco, Wortsman, Wightman, Gordon, Carlini, Taori, Dave, Shankar, Namkoong, Miller, Hajishirzi, Farhadi, and Schmidt]{ilharco_gabriel_2021_5143773}
Ilharco, G., Wortsman, M., Wightman, R., Gordon, C., Carlini, N., Taori, R., Dave, A., Shankar, V., Namkoong, H., Miller, J., Hajishirzi, H., Farhadi, A., and Schmidt, L.
\newblock Openclip, July 2021.
\newblock If you use this software, please cite it as below.

\bibitem[Li et~al.(2022)Li, Li, Xiong, and Hoi]{0001LXH22}
Li, J., Li, D., Xiong, C., and Hoi, S. C.~H.
\newblock {BLIP:} bootstrapping language-image pre-training for unified vision-language understanding and generation.
\newblock In Chaudhuri, K., Jegelka, S., Song, L., Szepesv{\'{a}}ri, C., Niu, G., and Sabato, S. (eds.), \emph{International Conference on Machine Learning, {ICML} 2022, 17-23 July 2022, Baltimore, Maryland, {USA}}, volume 162 of \emph{Proceedings of Machine Learning Research}, pp.\  12888--12900. {PMLR}, 2022.

\bibitem[Li et~al.(2023{\natexlab{a}})Li, Li, Savarese, and Hoi]{li2023blip2}
Li, J., Li, D., Savarese, S., and Hoi, S. C.~H.
\newblock {BLIP-2:} bootstrapping language-image pre-training with frozen image encoders and large language models.
\newblock In Krause, A., Brunskill, E., Cho, K., Engelhardt, B., Sabato, S., and Scarlett, J. (eds.), \emph{International Conference on Machine Learning, {ICML} 2023, 23-29 July 2023, Honolulu, Hawaii, {USA}}, volume 202 of \emph{Proceedings of Machine Learning Research}, pp.\  19730--19742. {PMLR}, 2023{\natexlab{a}}.

\bibitem[Li et~al.(2023{\natexlab{b}})Li, Wang, and Xie]{li2023inverse}
Li, X., Wang, Z., and Xie, C.
\newblock An inverse scaling law for {CLIP} training.
\newblock In Oh, A., Naumann, T., Globerson, A., Saenko, K., Hardt, M., and Levine, S. (eds.), \emph{Advances in Neural Information Processing Systems 36: Annual Conference on Neural Information Processing Systems 2023, NeurIPS 2023, New Orleans, LA, USA, December 10 - 16, 2023}, 2023{\natexlab{b}}.

\bibitem[Lin et~al.(2014)Lin, Maire, Belongie, Hays, Perona, Ramanan, Doll{\'{a}}r, and Zitnick]{lin2015microsoftcoco}
Lin, T., Maire, M., Belongie, S.~J., Hays, J., Perona, P., Ramanan, D., Doll{\'{a}}r, P., and Zitnick, C.~L.
\newblock Microsoft {COCO:} common objects in context.
\newblock In Fleet, D.~J., Pajdla, T., Schiele, B., and Tuytelaars, T. (eds.), \emph{Computer Vision - {ECCV} 2014 - 13th European Conference, Zurich, Switzerland, September 6-12, 2014, Proceedings, Part {V}}, volume 8693 of \emph{Lecture Notes in Computer Science}, pp.\  740--755. Springer, 2014.
\newblock \doi{10.1007/978-3-319-10602-1\_48}.

\bibitem[Liu et~al.(2023)Liu, Li, Wu, and Lee]{liu2023visual}
Liu, H., Li, C., Wu, Q., and Lee, Y.~J.
\newblock Visual instruction tuning.
\newblock In Oh, A., Naumann, T., Globerson, A., Saenko, K., Hardt, M., and Levine, S. (eds.), \emph{Advances in Neural Information Processing Systems 36: Annual Conference on Neural Information Processing Systems 2023, NeurIPS 2023, New Orleans, LA, USA, December 10 - 16, 2023}, 2023.

\bibitem[Liu et~al.(2021)Liu, Shen, Zhang, Dolan, Carin, and Chen]{liu2021makes}
Liu, J., Shen, D., Zhang, Y., Dolan, B., Carin, L., and Chen, W.
\newblock What makes good in-context examples for gpt-$3 $?
\newblock \emph{arXiv preprint arXiv:2101.06804}, 2021.

\bibitem[Min et~al.(2021)Min, Lewis, Zettlemoyer, and Hajishirzi]{min2021metaicl}
Min, S., Lewis, M., Zettlemoyer, L., and Hajishirzi, H.
\newblock Metaicl: Learning to learn in context.
\newblock \emph{arXiv preprint arXiv:2110.15943}, 2021.

\bibitem[Oquab et~al.(2024)Oquab, Darcet, Moutakanni, Vo, Szafraniec, Khalidov, Fernandez, Haziza, Massa, El-Nouby, Assran, Ballas, Galuba, Howes, Huang, Li, Misra, Rabbat, Sharma, Synnaeve, Xu, Jegou, Mairal, Labatut, Joulin, and Bojanowski]{oquab2024dinov2learningrobustvisual}
Oquab, M., Darcet, T., Moutakanni, T., Vo, H., Szafraniec, M., Khalidov, V., Fernandez, P., Haziza, D., Massa, F., El-Nouby, A., Assran, M., Ballas, N., Galuba, W., Howes, R., Huang, P.-Y., Li, S.-W., Misra, I., Rabbat, M., Sharma, V., Synnaeve, G., Xu, H., Jegou, H., Mairal, J., Labatut, P., Joulin, A., and Bojanowski, P.
\newblock Dinov2: Learning robust visual features without supervision, 2024.
\newblock URL \url{https://arxiv.org/abs/2304.07193}.

\bibitem[Radford et~al.(2018)Radford, Narasimhan, Salimans, Sutskever, et~al.]{radford2018improving}
Radford, A., Narasimhan, K., Salimans, T., Sutskever, I., et~al.
\newblock Improving language understanding by generative pre-training.
\newblock 2018.

\bibitem[Radford et~al.(2021)Radford, Kim, Hallacy, Ramesh, Goh, Agarwal, Sastry, Askell, Mishkin, Clark, Krueger, and Sutskever]{RadfordKHRGASAM21}
Radford, A., Kim, J.~W., Hallacy, C., Ramesh, A., Goh, G., Agarwal, S., Sastry, G., Askell, A., Mishkin, P., Clark, J., Krueger, G., and Sutskever, I.
\newblock Learning transferable visual models from natural language supervision.
\newblock In Meila, M. and Zhang, T. (eds.), \emph{Proceedings of the 38th International Conference on Machine Learning, {ICML} 2021, 18-24 July 2021, Virtual Event}, volume 139 of \emph{Proceedings of Machine Learning Research}, pp.\  8748--8763. {PMLR}, 2021.

\bibitem[Sehgal et~al.(2023)Sehgal, Grayeli, Sun, and Chaudhuri]{sehgal2024neurosymbolic}
Sehgal, A., Grayeli, A., Sun, J.~J., and Chaudhuri, S.
\newblock Neurosymbolic grounding for compositional world models.
\newblock \emph{CoRR}, abs/2310.12690, 2023.
\newblock \doi{10.48550/ARXIV.2310.12690}.

\bibitem[Singh et~al.(2022)Singh, Hu, Goswami, Couairon, Galuba, Rohrbach, and Kiela]{SinghHGCGRK22}
Singh, A., Hu, R., Goswami, V., Couairon, G., Galuba, W., Rohrbach, M., and Kiela, D.
\newblock {FLAVA:} {A} foundational language and vision alignment model.
\newblock In \emph{{IEEE/CVF} Conference on Computer Vision and Pattern Recognition, {CVPR} 2022, New Orleans, LA, USA, June 18-24, 2022}, pp.\  15617--15629. {IEEE}, 2022.
\newblock \doi{10.1109/CVPR52688.2022.01519}.

\bibitem[Sun et~al.(2023)Sun, Fang, Wu, Wang, and Cao]{sun2023evaclip}
Sun, Q., Fang, Y., Wu, L., Wang, X., and Cao, Y.
\newblock {EVA-CLIP:} improved training techniques for {CLIP} at scale.
\newblock \emph{CoRR}, abs/2303.15389, 2023.
\newblock \doi{10.48550/ARXIV.2303.15389}.

\bibitem[Thrush et~al.(2022)Thrush, Jiang, Bartolo, Singh, Williams, Kiela, and Ross]{thrush2022winoground}
Thrush, T., Jiang, R., Bartolo, M., Singh, A., Williams, A., Kiela, D., and Ross, C.
\newblock Winoground: Probing vision and language models for visio-linguistic compositionality.
\newblock In \emph{{IEEE/CVF} Conference on Computer Vision and Pattern Recognition, {CVPR} 2022, New Orleans, LA, USA, June 18-24, 2022}, pp.\  5228--5238. {IEEE}, 2022.
\newblock \doi{10.1109/CVPR52688.2022.00517}.

\bibitem[Touvron et~al.(2023)Touvron, Martin, Stone, Albert, Almahairi, Babaei, Bashlykov, Batra, Bhargava, Bhosale, Bikel, Blecher, Canton{-}Ferrer, Chen, Cucurull, Esiobu, Fernandes, Fu, Fu, Fuller, Gao, Goswami, Goyal, Hartshorn, Hosseini, Hou, Inan, Kardas, Kerkez, Khabsa, Kloumann, Korenev, Koura, Lachaux, Lavril, Lee, Liskovich, Lu, Mao, Martinet, Mihaylov, Mishra, Molybog, Nie, Poulton, Reizenstein, Rungta, Saladi, Schelten, Silva, Smith, Subramanian, Tan, Tang, Taylor, Williams, Kuan, Xu, Yan, Zarov, Zhang, Fan, Kambadur, Narang, Rodriguez, Stojnic, Edunov, and Scialom]{touvron2023llama}
Touvron, H., Martin, L., Stone, K., Albert, P., Almahairi, A., Babaei, Y., Bashlykov, N., Batra, S., Bhargava, P., Bhosale, S., Bikel, D., Blecher, L., Canton{-}Ferrer, C., Chen, M., Cucurull, G., Esiobu, D., Fernandes, J., Fu, J., Fu, W., Fuller, B., Gao, C., Goswami, V., Goyal, N., Hartshorn, A., Hosseini, S., Hou, R., Inan, H., Kardas, M., Kerkez, V., Khabsa, M., Kloumann, I., Korenev, A., Koura, P.~S., Lachaux, M., Lavril, T., Lee, J., Liskovich, D., Lu, Y., Mao, Y., Martinet, X., Mihaylov, T., Mishra, P., Molybog, I., Nie, Y., Poulton, A., Reizenstein, J., Rungta, R., Saladi, K., Schelten, A., Silva, R., Smith, E.~M., Subramanian, R., Tan, X.~E., Tang, B., Taylor, R., Williams, A., Kuan, J.~X., Xu, P., Yan, Z., Zarov, I., Zhang, Y., Fan, A., Kambadur, M., Narang, S., Rodriguez, A., Stojnic, R., Edunov, S., and Scialom, T.
\newblock Llama 2: Open foundation and fine-tuned chat models.
\newblock \emph{CoRR}, abs/2307.09288, 2023.
\newblock \doi{10.48550/ARXIV.2307.09288}.

\bibitem[Urbanek et~al.(2023)Urbanek, Bordes, Astolfi, Williamson, Sharma, and Romero{-}Soriano]{urbanek2023picture}
Urbanek, J., Bordes, F., Astolfi, P., Williamson, M., Sharma, V., and Romero{-}Soriano, A.
\newblock A picture is worth more than 77 text tokens: Evaluating clip-style models on dense captions.
\newblock \emph{CoRR}, abs/2312.08578, 2023.
\newblock \doi{10.48550/ARXIV.2312.08578}.

\bibitem[Urbanek et~al.(2024)Urbanek, Bordes, Astolfi, Williamson, Sharma, and Romero-Soriano]{urbanek2024pictureworth77text}
Urbanek, J., Bordes, F., Astolfi, P., Williamson, M., Sharma, V., and Romero-Soriano, A.
\newblock A picture is worth more than 77 text tokens: Evaluating clip-style models on dense captions, 2024.
\newblock URL \url{https://arxiv.org/abs/2312.08578}.

\bibitem[Wang et~al.(2023)Wang, Lv, Yu, Hong, Qi, Wang, Ji, Yang, Zhao, Song, Xu, Xu, Li, Dong, Ding, and Tang]{wang2024cogvlm}
Wang, W., Lv, Q., Yu, W., Hong, W., Qi, J., Wang, Y., Ji, J., Yang, Z., Zhao, L., Song, X., Xu, J., Xu, B., Li, J., Dong, Y., Ding, M., and Tang, J.
\newblock Cogvlm: Visual expert for pretrained language models.
\newblock \emph{CoRR}, abs/2311.03079, 2023.
\newblock \doi{10.48550/ARXIV.2311.03079}.

\bibitem[Wu et~al.(2022)Wu, Wang, Ye, and Kong]{wu2022self}
Wu, Z., Wang, Y., Ye, J., and Kong, L.
\newblock Self-adaptive in-context learning: An information compression perspective for in-context example selection and ordering.
\newblock \emph{arXiv preprint arXiv:2212.10375}, 2022.

\bibitem[Yu et~al.(2022)Yu, Wang, Vasudevan, Yeung, Seyedhosseini, and Wu]{yu2022coca}
Yu, J., Wang, Z., Vasudevan, V., Yeung, L., Seyedhosseini, M., and Wu, Y.
\newblock Coca: Contrastive captioners are image-text foundation models.
\newblock \emph{Trans. Mach. Learn. Res.}, 2022, 2022.

\bibitem[Y{\"{u}}ksekg{\"{o}}n{\"{u}}l et~al.(2023)Y{\"{u}}ksekg{\"{o}}n{\"{u}}l, Bianchi, Kalluri, Jurafsky, and Zou]{yuksekgonul2023visionlanguage}
Y{\"{u}}ksekg{\"{o}}n{\"{u}}l, M., Bianchi, F., Kalluri, P., Jurafsky, D., and Zou, J.
\newblock When and why vision-language models behave like bags-of-words, and what to do about it?
\newblock In \emph{The Eleventh International Conference on Learning Representations, {ICLR} 2023, Kigali, Rwanda, May 1-5, 2023}. OpenReview.net, 2023.

\bibitem[Yun et~al.(2022)Yun, Lee, Kim, and Shin]{yun2022patchpred}
Yun, S., Lee, H., Kim, J., and Shin, J.
\newblock Patch-level representation learning for self-supervised vision transformers, 2022.
\newblock URL \url{https://arxiv.org/abs/2206.07990}.

\bibitem[Zhai et~al.(2023)Zhai, Mustafa, Kolesnikov, and Beyer]{zhai2023sigmoid}
Zhai, X., Mustafa, B., Kolesnikov, A., and Beyer, L.
\newblock Sigmoid loss for language image pre-training.
\newblock In \emph{{IEEE/CVF} International Conference on Computer Vision, {ICCV} 2023, Paris, France, October 1-6, 2023}, pp.\  11941--11952. {IEEE}, 2023.
\newblock \doi{10.1109/ICCV51070.2023.01100}.

\end{thebibliography}

\bibliographystyle{uvafomo2024}

%%%%%%%%%%%%%%%%%%%%%%%%%%%%%%%%%%%%%%%%%%%%%%%%%%%%%%%%%%%%%%%%%%%%%%%%%%%%%%%
%%%%%%%%%%%%%%%%%%%%%%%%%%%%%%%%%%%%%%%%%%%%%%%%%%%%%%%%%%%%%%%%%%%%%%%%%%%%%%%
% APPENDIX
%%%%%%%%%%%%%%%%%%%%%%%%%%%%%%%%%%%%%%%%%%%%%%%%%%%%%%%%%%%%%%%%%%%%%%%%%%%%%%%
%%%%%%%%%%%%%%%%%%%%%%%%%%%%%%%%%%%%%%%%%%%%%%%%%%%%%%%%%%%%%%%%%%%%%%%%%%%%%%%
\newpage
\appendix
\onecolumn

\counterwithin{figure}{section}
\counterwithin{table}{section}
\counterwithin{equation}{section}

% \appendix

\section{Introduction} \label{app: intro}

\begin{figure}[h]
\begin{center}
\includegraphics[width=0.45\textwidth]{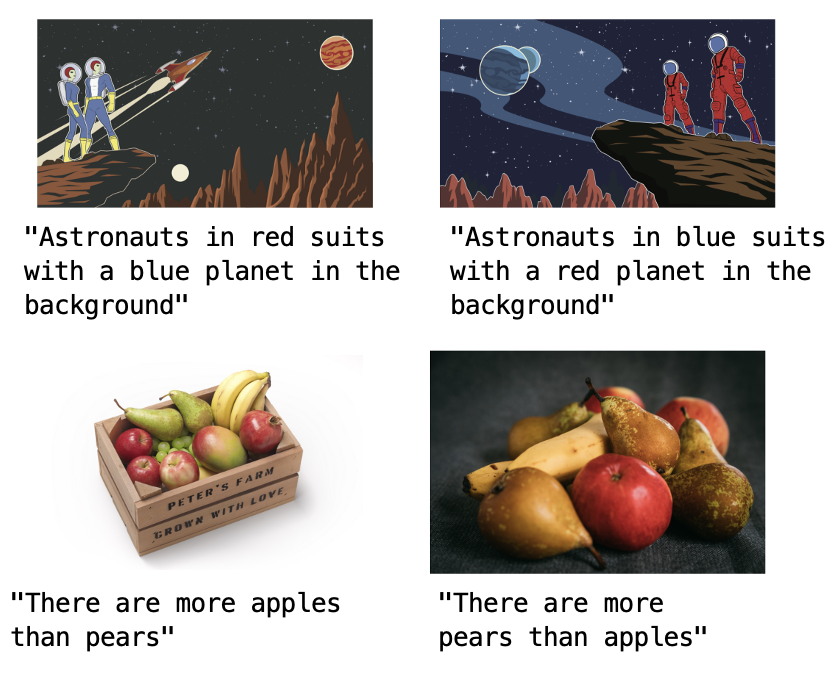}
\end{center}
\caption{\textbf{Compositional reasoning examples} from Winoground \cite{thrush2022winoground}, showing the close similarity between the pairs of images and text.}
\label{fig:winoground_example1}
\end{figure}

\section{Method} \label{app: method}
Below we state the ICL prompting strategy used in our experiments, 

\textit{USER: Does the image match the caption?\\A. $<$CaptionA$>$\\B. $<$CaptionB$>$\\$<$image1$>$. The correct caption is: A/B\\.\\. (We repeat the above 5 times for 5-shot in-context learning)\\.\\USER: Similarly, given an image and two captions choose the correct caption. Think step-by-step and analyze the captions against the image. Begin by describing the key elements visible in the image. Then, compare these elements with the details mentioned in the captions. Clearly state your final answer only in a single character, either A or B.\\$<$image$>$. The caption is: \\A. $<$CaptionA$>$\\B. $<$CaptionB$>$\\ASSISTANT:}

The prompting strategy used to generate the \textit{wrong} caption corresponding to the correct one using GPT-4o is as below,

\textit{Generate counter caption to this one, with the same objects in a different position/attribute: `correct caption'.}

% \begin{figure*}[h]
% \begin{center}
% \includegraphics[width=1\textwidth]{figures/WhatsApp Image 2024-05-28 at 15.51.24.jpeg}
% \end{center}
% \caption{\textbf{Few-shot samples}. \emph{First row}: The images and captions are synthetically generated with GPT-4o as seen in Figure \ref{fig:image_gen_pipeline} \emph{Second row}: Images are manually captioned and retrieved from COCO dataset \cite{lin2015microsoftcoco}. We use these images to instill an understanding of the task within the generative models in a few-shot manner.}
% \label{fig:few_shot_examples}
% \end{figure*}

\section{Appendix Experiments} \label{app: experiments}
The ICL prompting strategy used in SugarCrepe and ARO evaluation is as follows, 

\textit{USER: $<$image$>$ Given this image and two candidate captions (A and B), which caption is the better description of the given image? Clearly state your final answer only in a single character, either A or B.\\A. $<$CaptionA$>$\\B. $<$CaptionB$>$}

The ICL prompting strategy used in Winoground evaluation is as follows, 

\textit{After providing a brief explanation of your reasoning, clearly state your final answer as $<$Yes$>$ or $<$No$>$.}

\begin{table}[t]
    \centering
    \setlength\aboverulesep{0pt}
    \setlength\belowrulesep{0pt}
    \setstretch{1.2}
    \begin{tabular}{lll}
    \toprule
    \textbf{Model}   & \textbf{Vision encoder}    & \textbf{LLM} \\
    \midrule
    LLaVA   & CLIP ViT-L-336px  & Vicuna1.5-7B  \\
    CogVLM  & EVA-02-CLIP       & Vicuna1.5-7B \\
    \bottomrule
    \end{tabular}
    \caption{\textbf{Generative VLMs} and the vision encoders and LLMs they use}
    \label{tab:generative_models}
\end{table}

\section{Pipeline details} \label{app: pipeline}

\subsection{Contrastive evaluation pipeline}
% In this section we describe the evaluation pipeline employed in our experiments for contrastive models.

\paragraph{ARO and SugarCrepe}

For contrastive models, we evaluate ARO and SugarCrepe by first taking the positive and negative captions embeddings and comparing both with the embeddings of the image. We do this by computing the cosine similarity between each caption and the image embedding and increasing the number of correct predictions when the positive caption-image score is higher than the negative caption-image score. We adapt the code \footnote{\url{https://github.com/mertyg/vision-language-models-are-bows/blob/main/model_zoo/clip_models.py}} 
by \cite{yuksekgonul2023visionlanguage}.

\paragraph{Winoground}

For the Winoground benchmark, we follow \cite{ilharco_gabriel_2021_5143773} and perform a text and image encoding, for each image-caption pair. This results in two image feature representations and two caption feature representations. The final scores are then calculated by taking the cosine similarity score between the representations. This returns the real-valued outputs, which are then used to determine the text-image-group scores.

% \mn{Do we talk about the other evaluation method for generative models? I mean the one without logits? I kinda did in \ref{sec: logits_method}}

\subsection{Generative evaluation pipeline}

\paragraph{ARO and SugarCrepe}

For generative models, we evaluate ARO and SugarCrepe zero-shot by prompting the models using the ICL method shown in Appendix \ref{app: experiments}.
We then check the output of the model and increase the number of correct predictions if the model picks the correct caption choice. For 1-shot and 5-shot in-context learning, we use the prompt mentioned in Appendix \ref{app: method}.

\paragraph{Winoground}

For the Winoground benchmark, we use a separate final instruction in the previous prompts as stated in Appendix \ref{app: experiments}. 
If a \textit{``yes"} character is found in the output, then the result of that corresponding pair is set to 1, if not it is set to 0. 
However, this evaluation strategy causes two major issues. 
First, the output is not always the same. Variations in the outputs result in both categorizing a correct caption as wrong if \textit{``yes"} is never predicted and vice-versa if the predicted \textit{``yes"} is not relating to the caption entailing the image/or choice but rather something else.
To quantify this, consider the probability distribution $P(t)$ of token $t \in V$ (Vocabulary) across the sequence length $s$, derived from the logits $L$ using the softmax function.
% \begin{equation}
%     P(t) = \frac{e^{L_{s, t}}}{\sum_{i=1}^{V} e^{L_{s, i}}}.
% \end{equation}
Even if $P(t)$ is high, $t$ might not be generated if another token has a higher probability.
Secondly, given the binary value of 0/1, evaluating generative models on Winoground using the previous method results in having \textit{text, image, group scores} to be all equal. 
To mitigate the aforementioned issues, we propose an alternative that relies on using the output logits of the desired word for evaluation. In this method, we first take the logits output tensor $L \in \mathbb{R}^{B \times S \times V} $, where $B$ is the batch size (equal to 1 in this instance), $S$ is sequence length and $V$ is the vocabulary size. We take the token id \textit{"yes"} (denoted as $id_{yes}$) in the third dimension, and compute the mean over the sequence length, $L_{yes} = \frac{1}{S} \sum_{s=1}^{S} L_{s, id_{yes}}$.

This results in a real-valued number $L_{yes} \in \mathbb{R}$, one per each caption-image pair given as input to the model. These values will then be compared in the same way as we do in contrastive evaluation to obtain the three accuracy scores.
This technique is beneficial over the first one because it does not directly rely on generation, rather it focuses on the amount of ``confidence" the model had about a specific token throughout the whole generated sequence.

% The $\mathtt{\backslash onecolumn}$ command above can be kept in place if you prefer a one-column appendix, or can be removed if you prefer a two-column appendix.  Apart from this possible change, the style (font size, spacing, margins, page numbering, etc.) should be kept the same as the main body.
%%%%%%%%%%%%%%%%%%%%%%%%%%%%%%%%%%%%%%%%%%%%%%%%%%%%%%%%%%%%%%%%%%%%%%%%%%%%%%%
%%%%%%%%%%%%%%%%%%%%%%%%%%%%%%%%%%%%%%%%%%%%%%%%%%%%%%%%%%%%%%%%%%%%%%%%%%%%%%%

\end{document}